\documentclass{bmvc2k}

\usepackage{amsmath}
\usepackage{amssymb}
\usepackage{enumitem}
\usepackage{xcolor,microtype}
\usepackage{dsfont}

\title{Cluster-level pseudo-labelling for source-free cross-domain facial expression recognition}

\addauthor{Alessandro Conti}{alessandro.conti-1@unitn.it}{1}
\addauthor{Paolo Rota}{paolo.rota@unitn.it}{1,2}
\addauthor{Yiming Wang}{ywang@fbk.eu}{3}
\addauthor{Elisa Ricci}{e.ricci@unitn.it}{1,3}

\addinstitution{
 DISI - Department of Information Engineering and Computer Science\\
 University of Trento\\
}

\addinstitution{
 CIMeC - Center for Mind and Brain Sciences\\
 University of Trento\\
}

\addinstitution{
 Fondazione Bruno Kessler (FBK)
}

\runninghead{Conti, Rota, Wang, Ricci}{Cluster-level pseudo-labelling}

\newcommand{\mname}[0]{CluP}
\newcommand{\ie}{\textit{i}.\textit{e}. }

\newcommand{\data}{\mathcal{D}}
\newcommand{\tsource}{\mathtt{S}}
\newcommand{\ttarget}{\mathtt{T}}

\begin{document}

\maketitle

\begin{abstract}
Automatically understanding emotions from visual data is a fundamental task for human behaviour understanding. While models devised for Facial Expression Recognition (FER) have demonstrated excellent performances on many datasets, they often suffer from severe performance degradation when trained and tested on different datasets due to domain shift. In addition, as face images are considered highly sensitive data, the accessibility to large-scale datasets for model training is often denied. In this work, we tackle the above-mentioned problems by proposing the first Source-Free Unsupervised Domain Adaptation (SFUDA) method for FER. Our method exploits self-supervised pretraining to learn good feature representations from the target data and proposes a novel and robust cluster-level pseudo-labelling strategy that accounts for in-cluster statistics. We validate the effectiveness of our method in four adaptation setups, proving that it consistently outperforms existing SFUDA methods when applied to FER, and is on par with methods addressing FER in the UDA setting.

\noindent Code is available at \url{https://github.com/altndrr/clup}.
\end{abstract}

%-------------------------------------------------------------------------
\section{Introduction}
\label{sec:intro}
\begin{figure}[t]
\begin{center}
\includegraphics[width=0.9\linewidth]{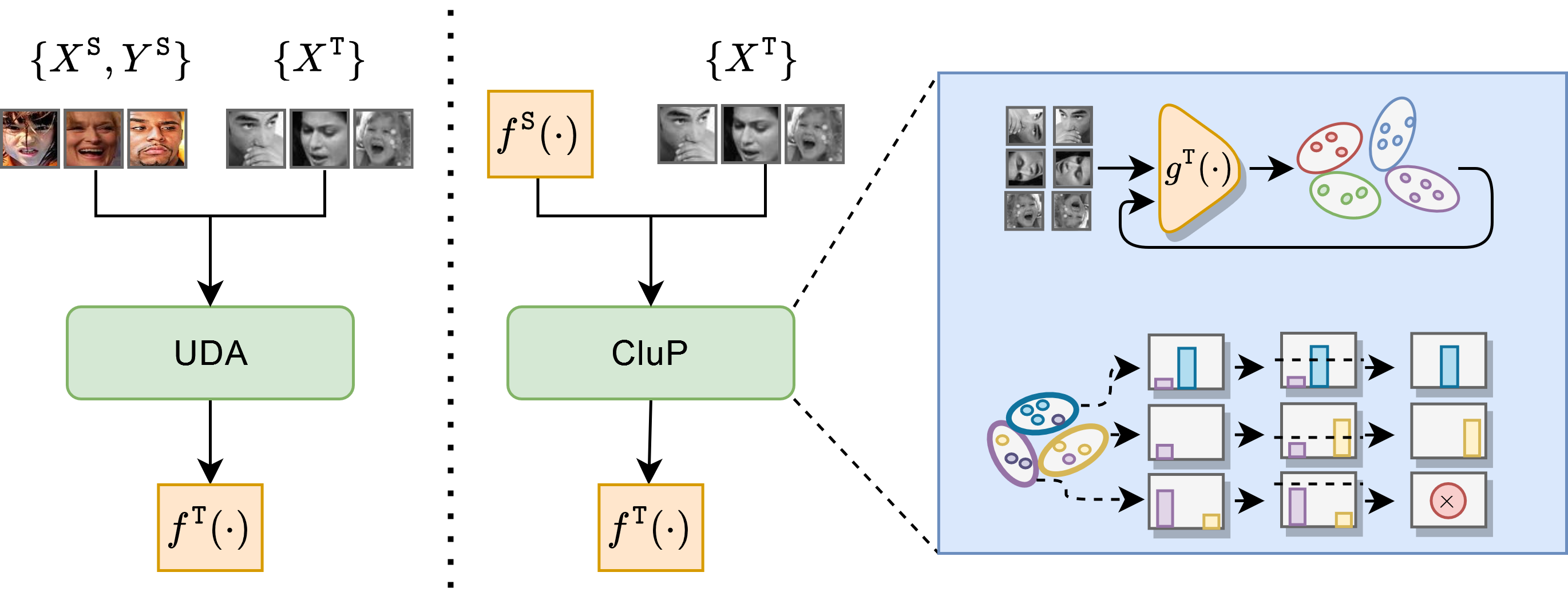}
\end{center}
   \caption{Comparison between previous works (the left part) and our \mname~on cross-domain FER (the right part).
   Differently from past works, we aim to learn a target model $f^\ttarget(\cdot)$ with only source model $f^\tsource(\cdot)$ and unlabelled target data $\left\{X^\ttarget\right\}$ without the source data $\left\{X^\tsource,~Y^\tsource\right\}$, a very likely scenario due to privacy concerns. Our solution, \mname, is the first method on source-free domain adaptation for FER, exploiting self-supervised learning (SSL) to warm up the target feature extractor $g^\ttarget(\cdot)$ and a novel cluster-level pseudo-labelling technique.}
\label{fig:intro:teaser}
\end{figure}

Facial Expression Recognition (FER)~\cite{shi2021learning, savchenko2021facial, wang2020region, hasani2020breg} refers to the task of automatically inferring the emotional state of a person from a facial image, which supports multiple application fields, such as assistive robotics and security monitoring.
However, each individual shows their emotional state differently according to their personal traits or complicated cultural/ethical factors~\cite{calvo2016perceptual}. Such heterogeneity in the data space remains one of the main challenges for a generalisable model for FER.
In the last twenty years, the efforts to improve such technologies have been mostly split between collecting larger and more diverse datasets~\cite{zafeiriou2017aff,mollahosseini2017affectnet} and advancing learning algorithms for improving generalisation capability in the wild \cite{zhang2021relative, hasani2020breg}.
Many recent techniques for FER exploit the attention mechanism~\cite{minaee2021deep, wen2021distract, aouayeb2021learning, pecoraro2021local, farzaneh2021facial}, while some other works learn uncertainty via feature mixup~\cite{zhang2021relative}, or improve feature representations by replacing the pooling layers to reduce padding erosion~\cite{shi2021learning}.

Recent works often frame the problem from an Unsupervised Domain Adaptation (UDA) perspective where labels of the target samples are not available~\cite{ji2019cross,li2018deep,li2020deeper}.
For example, in~\cite{li2017reliable}, Li \textit{et al.} introduce a novel loss function to preserve feature locality despite the domain shift.
Such loss also organises facial expressions according to their intensity in the embedding space.
A more recent method~\cite{chen2021cross} exploits facial landmarks and holistic features to adapt to the target domain with adversarial learning applied on graphs.

While all these methods improve the adaptability of FER models across data distributions, the source data is required during adaptation.
However, when dealing with facial images, the source data might not be available due to the increasingly stringent regulations concerning the privacy of citizens. 
Therefore, we are motivated to address the more challenging problem of Source-Free Unsupervised Domain Adaptation (SFUDA) for FER, given only the availability of the source pretrained model (see Fig.~\ref{fig:intro:teaser}).
To the best of our knowledge, we are the first to propose a domain adaptation solution for FER that works without the source facial data, embracing a privacy-preserving learning paradigm as the source data can remain private.

Our proposed method, \mname~(\textbf{Clu}ster-level \textbf{P}seudo-labelling), exploits self-supervised learning (SSL) on the target data and proposes a novel cluster-level pseudo-labelling technique.
Pseudo-labelling for UDA often extends the source model to the target domain using the source confidence to select the best target training inputs~\cite{zhang2018collaborative,long2017deep}.
However, the computation of confidence requires supervised training, which is only possible in UDA with the access to the source data.
In the case of SFUDA, as the domain gap increases, one can expect a degrading representation capability of the source model on the target domain. Recent advances in SSL shows that a good data representation can be learnt without annotated labels~\cite{caron2020unsupervised,chen2020simple,he2020momentum}. 
In this work, we propose to exploit SSL techniques for a good starting feature representation for the target model, and further propose to improve the reliability of pseudo-labels with our newly introduced \textit{cluster purity}, \ie the local statistics of target samples that are clustered within the feature space expressed by the source model.
We validate \mname\ on a set of cross-domain FER benchmarks and prove its advantageous performance in terms of classification accuracy.

We summarise our contributions as follows:
\begin{itemize}[noitemsep,nolistsep]
    \item We present \mname, the first method addressing Source-free Unsupervised Domain Adaptation for FER, exploiting SSL to foundation our target model.
    \item \mname\ introduces a novel cluster-level pseudo-labelling scheme to improve the reliability of pseudo-labels based on in-cluster attributes that deviates from traditional confidence-based pseudo-labelling methods.
    \item We demonstrate that \mname\ surpasses competing methods for SFUDA and is comparable with UDA techniques on several FER adaptation benchmarks.
\end{itemize}

%%%%%%%%%
\section{Related work}
\label{sec:sota}

In the following, we present recent works on UDA methods for FER, and some general-purpose SFUDA solutions.

\noindent\textbf{Unsupervised Domain Adaptation for FER.}
As a consequence of the domain bias, quite prominent among FER datasets, some works focus on domain adaptation~\cite{ji2019cross, zhu2016discriminative, li2017reliable, li2018deep, zavarez2017cross, li2020deeper, chen2021cross}.
In~\cite{ji2019cross}, Ji \textit{et al.} apply late fusion on the outputs of two channels that learn intra-category and inter-category similarities of facial expressions.
The authors of~\cite{li2017reliable} introduce a locality preserving loss that draws samples of the same class closer.
They also notice that neighbouring samples in the embedding space present similar emotional intensities.
DETN~\cite{li2018deep} applies two variations of the Maximum Mean Discrepancy to assess the amount of divergence between the domains and to re-weight the class-wise source distribution to match the target.
The authors extend the work in~\cite{li2020deeper}, where they additionally consider the differences in the conditional distributions.
Differently from the above works, AGRA~\cite{chen2021cross} focuses on the well-established approach of adversarial domain adaptation, leveraging facial landmarks alongside facial images.
For the landmarks, they introduce two specialised graph neural networks while jointly considering the domain feature distributions, the local features (\ie, landmarks), and the holistic features, achieving the best results on many benchmarks.

Compared to previous works, we consider a stricter setting where the source data is unavailable.
We argue that, due to privacy issues, human behaviour understanding methods do not always have access to the source data.
For this reason, we introduce a novel method for FER that adapt to a target domain in a source-free manner.

\noindent\textbf{Source-Free Unsupervised Domain Adaptation.}
Recently, novel methods for source-free domain adaptation have been proposed~\cite{liang2020we, hou2021visualizing, kurmi2021domain, li2020model, zhang2021unsupervised, liang2022dine}.
The setting represents a more complex but realistic scenario of UDA, where source data is unavailable.
Some works resort to entropy-minimisation losses to adapt to the target domain without labels.
For example, SHOT~\cite{liang2020we} employs an entropy loss alongside a classification loss on pseudo-labelled samples to adapt the network to the target domain.
The work has been extended in~\cite{liang2021source} introducing an auxiliary head that solves relative rotation, leading to improved performance.
Differently from the above, the authors of~\cite{hou2021visualizing} frame the problem from an image translation perspective and translate the target images to the source style using only the source model.
In~\cite{yang2021exploiting}, they perform self-training with a loss function that considers the intrinsic structure of the target domain via nearest neighbours.
In the proposed work, we do not impose any constraint on the loss function, our refinement step works on source clusters, and we propose a novel score function to select the best samples to train on the target domain.
Other works address open-set or universal domain adaptation~\cite{kundu2020towards, kundu2020universal} without access to the source data.

Unlike the previous works, our model does not rely only on the source model but is constructed based on independent self-supervised training on the target data. Moreover, we refine pseudo-labels by reducing unreliable samples using a novel decision metric at the cluster level based on cluster purity.

%%%%%%%%% 
\section{Method}
\label{sec:method}

The traditional closed-set UDA problem setting allows the access to the annotated source dataset $\data^\tsource=\{\mathbf{x}^\tsource_i, y^\tsource_i\}_{i=1}^{M^\tsource}$, and a target dataset $\data^\ttarget=\{\mathbf{x}^\ttarget_i\}_{i=1}^{M^\ttarget}$ without annotations, where the target domain shares the same label space with the source, \ie $\mathcal{Y}^S = \mathcal{Y}^\ttarget =\{1,~...,~N\}$.

Differently, the SFUDA protocol does not allow the access to the source dataset $\data^\tsource$, but only to a trained source model $f^\tsource(\cdot): \mathcal{X}^\tsource \rightarrow \mathbb{R}^N$, which consists of a feature extractor $g^\tsource(\cdot): \mathcal{X}^\tsource \rightarrow \mathbb{R}^Z$ and a classifier $h^\tsource(\cdot): \mathbb{R}^Z \rightarrow \mathbb{R}^N$, where $Z$ is the feature dimension.

\begin{figure*}
\begin{center}
\includegraphics[width=0.9\linewidth]{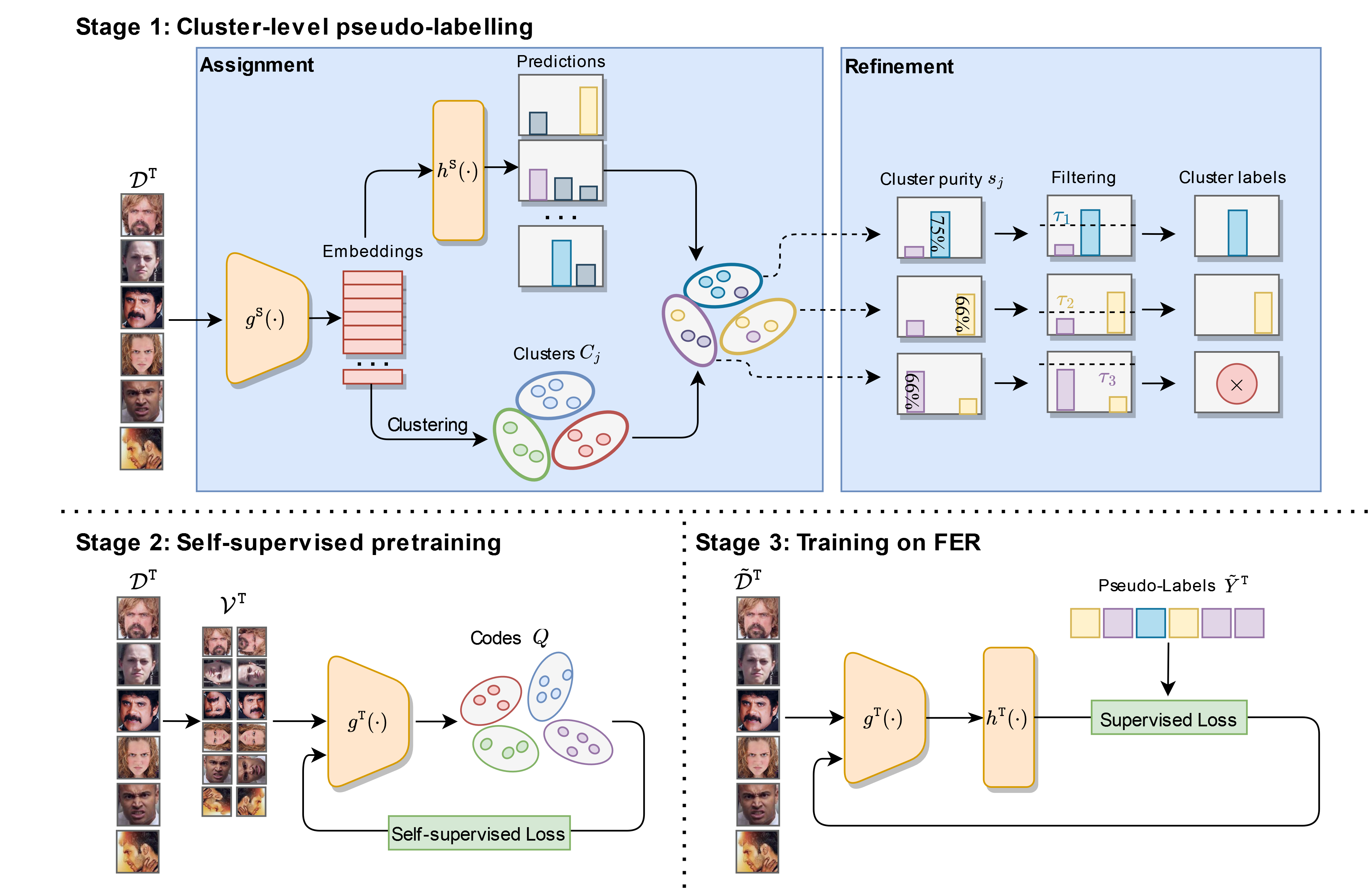}
\end{center}
   \caption{Our proposed \mname\ comprises of three-stage training, where the first stage produces trustworthy cluster-level pseudo-labels using the source model, the second stage warms up the target model in a self-supervised fashion, and finally the third stage performs the target model training with the refined pseudo-labels.
}
\label{fig:method:pipeline}
\end{figure*}

Our proposed method \mname\ tackles the problem of SFUDA for FER. As illustrated in Fig.~\ref{fig:method:pipeline}, \mname\ follows a three-stage training strategy where the first two can run in parallel: the first stage produces more trustworthy cluster-level pseudo-labels $\left\{\tilde{y}^\ttarget_i\right\}_{i=1}^{\tilde{M}^\ttarget}$ for a subset of $\tilde{M}^\ttarget$ target samples $\tilde{\data}^\ttarget=\{\mathbf{x}^\ttarget_i\}_{i=1}^{\tilde{M}^\ttarget}$ by exploiting the available $f^\tsource(\cdot)$ and our proposed cluster purity for pseudo-label refinement (described in Sec.~\ref{sec:method:pseudo-label-refinement}), while in the second stage, a target feature extractor $g^\ttarget(\cdot)$ is learned in a self-supervised fashion (described in Sec.~\ref{sec:method:pretraining}).
During the third stage, $g^\ttarget(\cdot)$ is extended with a classifier $h^\ttarget(\cdot)$ and the whole network is trained with the subset of target samples $\tilde{\data}^\ttarget$ accompanied by their refined pseudo-labels (described in Sec.~\ref{sec:method:downstream}).

\subsection{Cluster-level Pseudo-labelling}
\label{sec:method:pseudo-label-refinement}
Pseudo-labels filtered by confidence that is produced by source model are often unreliable, particularly when the domain gap between the source and target is large. \mname\ exploits a clustering technique to group samples with similar characteristics (i.e~\textit{assignment}) and then uses a purity metric based on the source classifier to select the most reliable clusters (\ie~\textit{refinement}). 

\noindent \textbf{Cluster pseudo-label assignment.} First, the target features are extracted $\left\{\mathbf{z}^\ttarget_i\right\}_{i=1}^{M^\ttarget} \in \mathbb{R}^Z$ 
using the source feature extractor $g^\tsource(\cdot)$. Second, we cluster the features using $K$-means algorithm, resulting in a set of clusters $\left\{C_j\right\}_{j=1}^K$
Since FER often deals with highly unbalanced datasets, we perform over-clustering and consider $K \gg N$, to increase the chances that even minor classes can be expressed with some clusters. 
Leveraging the pseudo-labels predicted by the source model $\tilde{y}^\ttarget_i=h^\tsource(\mathbf{z}^\ttarget_i)$ we assign to each cluster $C_j$ a pseudo-label $\tilde{y}_j^{\ttarget}$ that represents the majority-voted pseudo-label within each cluster.

\noindent \textbf{Cluster pseudo-label refinement.} As each cluster $C_j$ should contain elements that are similar in the learned feature space, we might expect their pseudo-labels to expose an one-class distribution.
Unfortunately, this is often not the case. However, a subset of clusters detaining a certain pseudo-label agreement can be defined using what we named as \textit{cluster purity}. 

Let us consider $m_j^\ttarget$ as the cardinality of the $j$-th cluster, where $M^\ttarget = \sum_{j=1}^{K} m_j^\ttarget$.
We define the \textit{cluster purity} score ${s}_{j}$ for each cluster $C_j$ as the percentage of pseudo-labels $\left\{ \tilde{y}_{i}^{\ttarget} \right\}_{i=1}^{m_j^T}$ that agree with their cluster-level label $\tilde{y}_{j}^{\ttarget}$:
\begin{equation}
 s_{j}  =  \frac{\sum_{i=1}^{m_j^T}\mathds{1}\left\{\textbf{x}^\ttarget_i \in C_j: \tilde{y}_i^{\ttarget} = \tilde{y}_{j}^{\ttarget}\right\}}{m_j^T}.
\end{equation}

Given $s_{j}$ per cluster, we can further refine the target dataset by only keeping clusters that have a \textit{cluster purity} score higher than a threshold $\tau$, \ie the more reliable clusters, for training the target model. Considering that each category of the pseudo-labels might exhibit a different distribution, we design the \textit{cluster purity} threshold $\tau$ to vary according to its category.    
Specifically, for the set of clusters that correspond to the same pseudo-label category $\left\{ C_j\right\}_{ \tilde{y}_{j}^{\ttarget}=n}$ where $n \in \mathcal{Y}$, we select the $Q$-th percentile of their purity scores to serve as the threshold ${\tau}_{n}$. $Q$ is empirically set, and related experimental details are reported in Sec.~\ref{sec:exp:ablation}. 

After the cluster refinement, only clusters whose \textit{cluster purity} score is higher than the threshold corresponding its category remain in the reduced target dataset $\tilde{\data}^\ttarget$ and will be used for training the final target model $f^\ttarget(\cdot)$.

\subsection{Self-supervised pretraining}
\label{sec:method:pretraining}
The pretraining of the target model is a delicate and important phase where the choice of the training method for warming up the backbone leads to relevant fluctuations in performance (see Sec.~\ref{sec:exp:ablation}). In the specific, we noticed that a pretraining relying on self-supervision largely outperforms a model initialised with source weights. For this reason, inspired by SwAV~\cite{caron2020unsupervised}, \mname~exploits self-supervision on the target dataset to learn an initial feature extractor $g^\ttarget(\cdot)$.

\mname~performs clustering of the sample data while enforcing the consistency between cluster assignments produced for different augmentations of the same sample.
First, target features $\{g^\ttarget(\textbf{x}^\ttarget_i)\}_{i=1}^{M^\ttarget}$ are grouped according to a similarity metric to retrieve $N^P$ learnable prototypes $P = \{p_i\}_{i=1}^{N^P}$ and a set of codes $\{q_i^{\ttarget}\}_{i=1}^{M^\ttarget}$ where each sample is assigned to.
Then, codes $\{q_i^{\ttarget}\}_{i=1}^{M^\ttarget}$ are used as targets to learn the optimal mapping to $\{g^\ttarget(\textbf{x}^\ttarget_i)\}_{i=1}^{M^\ttarget}$ by minimising:

\begin{equation}
    \mathcal L_{c}(\mathbf{x}_i, \mathbf{q}_i)=-\sum_{n=1}^{N^P} \textbf{q}_i^{(n)} \log(\textbf{p}_i^{(n)})
\end{equation}

\noindent where $\textbf{q}$ is the one-hot vector of $q$ and $\textbf{p}$ is the softmax of the dot product of $g^\ttarget(\textbf{x}_i^\ttarget)$ and the cluster prototypes $P$.

By treating each sample as a class (\ie $M^\ttarget=N$), contrastive learning aims to learn a feature extractor $g^\ttarget(\cdot)$ invariant to data augmentations.
For each target image $\textbf{x}_i^\ttarget$, we generate an arbitrary number $N^J$ of ``views'' by means of augmentation, \ie $\{\textbf{v}_{ij}^\ttarget = t_j(\textbf{x}_i^\ttarget)\}{_{i=1}^{M^\ttarget}}{_{j=1}^{N^J}}$ with $t_j(\cdot)\sim\mathcal{T}$.
Features extracted from views $\{g^\ttarget(\textbf{v}_{ij}^\ttarget)\}{_{i=1}^{M^\ttarget}}{_{j=1}^{N^J}}$ instead of from inputs $\{g^\ttarget(\textbf{x}_i^\ttarget)\}_{i=1}^{M^\ttarget}$ are then clustered.
The feature extractor $g^\ttarget(\cdot)$ aims to optimise for a ``swapped'' assignment problem between pairs of views $(j, k)\in \{1, ..., N^J\}$ of the same input $i\in \{1, ..., M^\ttarget\}$:

\begin{equation}
\label{eq:swapped}
    \mathcal L_{swapped}((\mathbf{v}_{ij}, \mathbf{q}_{ij}), (\mathbf{v}_{ik}, \mathbf{q}_{ik})) = \mathcal L_{c}(\textbf{v}_{ij}, \textbf{q}_{ik}) + \mathcal L_{c}(\textbf{v}_{ik}, \textbf{q}_{ij})
\end{equation}

\noindent We minimise $\mathcal L_{swapped}$ for all the pairs generated from $\data^\ttarget$ to get our pretrained $g^\ttarget(\cdot)$.

Finally, the whole model is trained by alternating between clustering features and minimising Eq.~\eqref{eq:swapped}.
To work online, clustering is reformulated as an optimal transport problem (as in \cite{asano2019self}) and is applied only on the features in a batch.

\subsection{Training on FER}
\label{sec:method:downstream}

Finally, the target model (\ie$f^\ttarget(\cdot)$) is obtained training the pseudo-labelled subset $\tilde{\data}^\ttarget$ (as detailed in Sec.~\ref{sec:method:pseudo-label-refinement}) by using the self-supervised feature extractor $g^\ttarget(\cdot)$ (as detailed in Sec.~\ref{sec:method:pretraining}) and a new classifier $h^\ttarget(\cdot)$. The model is trained with supervised cross-entropy loss between $\{\tilde{\textbf{y}}_i^\ttarget\}_{i=1}^{\tilde{M}^\ttarget}$ and the prediction $\{f^{\ttarget}(\textbf{x}_i^{\ttarget}\}_{i=1}^{\tilde{M}^\ttarget}$ as in Eq.~\eqref{eq:finalce}:

\begin{equation}
\label{eq:finalce}
    \mathcal L_{CE}^T = -\frac{1}{\tilde{M}^\ttarget}\sum_{i=1}^{\tilde{M}^\ttarget} \tilde{\textbf{y}_i^\ttarget}~\log{f^\ttarget\left (\textbf{x}_i^\ttarget\right )} 
\end{equation}

\noindent where $f(\cdot)$ already includes a softmax function for normalising the network logits into a probability distribution.

\section{Experiments}
\label{sec:exp}

We compare our method against the state-of-the-art methods for cross-domain FER with a set of benchmark datasets. We first introduce our experimental setup and then present the main comparison, followed by an extensive ablation study to justify our design choices.  

\noindent\textbf{Datasets.}
Following~\cite{chen2021cross}, we use AFE~\cite{chen2021cross} and RAF-DB~\cite{li2017reliable} as our source datasets, and ExpW~\cite{zhang2018facial} and FER2013~\cite{goodfellow2013challenges} as the target datasets.

\begin{itemize}[noitemsep,nolistsep]
    \item \textbf{AFE}~\cite{chen2021cross} contains 54,901 images of thousands of Asian individuals, collected from Asian films. This dataset addresses cross-culture domain adaptation, as the other datasets in our experiment involve mainly European and American people.
    \item \textbf{RAF-DB}~\cite{li2017reliable} contains 29,672 facial images from thousands of individuals that were collected from the Internet. We use RAF-DB as one of our source domain as it works as a counterpart of AFE.
    \item \textbf{ExpW}~\cite{zhang2018facial} contains 91,793 faces downloaded from Google Images, representing a large scale in-the-wild scenario with diverse ethic groups and facial poses.
    \item \textbf{FER2013}~\cite{goodfellow2013challenges} is large-scale dataset collected with the Google Images Search API, containing 35,887 grey images of low resolution.
    We consider FER2013 as a target domain to demonstrate cross-colour domain adaptation.
\end{itemize}

\begin{table*}[t!]
\begin{center}
\resizebox{0.95\linewidth}{!}{
\begin{tabular}{|ccccc|c|}
\hline
Method                           & AFE → ExpW  & AFE → FER2013   & RAF-DB → ExpW & RAF-DB → FER2013  \\
\hline\hline
ICID \cite{ji2019cross}          & 54.85       & 46.44           & 68.52        & 53.00            \\
DFA \cite{zhu2016discriminative} & 62.53       & 36.88           & 47.42        & 47.88            \\
LPL \cite{li2017reliable}        & 54.51       & 49.82           & 68.35        & 53.61            \\
DETN \cite{li2018deep}           & 58.41       & 45.39           & 43.92        & 42.01            \\
FTDNN \cite{zavarez2017cross}    & 55.29       & 48.58           & 68.08        & 53.28            \\
ECAN \cite{li2020deeper}         & 62.52       & 46.15           & 48.73        & 50.76            \\
CADA \cite{long2018conditional}  & 58.50       & 48.61           & 63.74        & 54.71            \\
SAFN \cite{xu2019larger}         & 55.17       & 50.07           & 68.32        & 53.31            \\
SWD \cite{lee2019sliced}         & 56.56       & 51.84           & 65.85        & 53.70            \\
AGRA \cite{chen2021cross}        & \it{65.03}  & 51.95           & \it{69.70}   & \it{54.94}       \\
\hline\hline
SHOT-IM \cite{liang2020we}       & 53.52       & 49.51           & 53.13        & 49.44            \\
SHOT \cite{liang2020we}          & 54.12       & 49.44           & 53.51        & 49.36            \\
\mname\ (DeepClusterV2)          & 62.56       & 50.47           & 65.43        & \bf{53.83}       \\
\mname\ (SwAV)                   & \bf{65.00}  & \textbf{\textit{52.51}}  & \bf{66.60}   & 53.71   \\
\hline
\end{tabular}}
\end{center}
\caption{Results of different methods in four domain adaptation settings, where the upper part lists methods for FER in the UDA setting with the source data accessible, while the lower part lists methods for FER in the SFUDA setting without accessing the source data. 
We highlight in \textit{italic} the best result among all methods and in \textbf{bold} the best among SFUDA ones.
Note that in ``AFE → FER2013'', \mname~achieves the best result among all methods.}
\label{tab:exp:main_results}
\end{table*}

\begin{table*}[t!]
\begin{center}
\resizebox{0.95\linewidth}{!}{
\begin{tabular}{|cccccccc|c|}
\hline
Method                     & Surprise    & Fear       & Disgust    & Happiness  & Sadness    & Anger      & Neutral    \\
\hline\hline
SHOT-IM \cite{liang2020we} & 28.29       & \bf{45.05} & 9.86       & 75.97      & 56.12      & 40.66      & 71.96      \\
SHOT \cite{liang2020we}    & 28.18       & 43.24      & 10.25      & 75.59      & 53.53      & 40.18      & 74.37      \\
\mname\ (DeepClusterV2)    & 29.44       & \bf{45.05} & 2.83       & \bf{83.15} & \bf{77.70} & 34.72      & \bf{76.65} \\
\mname\ (SwAV)             & \bf{37.89}  & 30.63      & \bf{13.57} & 80.72      & 50.85      & \bf{44.51} & 74.49      \\
\hline
\end{tabular}}
\end{center}
\caption{Class-wise accuracy for RAF-DB → FER2013 in SFUDA setting.}
\label{tab:exp:classwise_results}
\end{table*}

\noindent\textbf{Performance metric.} To evaluate the performance of our method, we consider traditional top-1 classification accuracy.
In addition, we also provide class-wise accuracy in our ablations to demonstrate how our method performs on different classes.

\noindent\textbf{Implementation details.}
We implement our method using PyTorch and PyTorch Lightning, and run all the experiments on NVIDIA A100 GPUs. We pretrain ResNet18 for FER as our source model, while we perform the self-supervised learning on the target domain using the solo-learn library~\cite{da2022solo} for 1000 epochs with SGD and a cosine annealing scheduling policy.
When performing cluster-level pseudo-labelling, we consider a large number of clusters for K-means to address imbalanced datasets.
We consider $K=1000$ for AFE → ExpW and RAF-DB → FER2013, and $K=250$ for the others.
We set the $Q$-th percentile per class to threshold the cluster purity, where $Q$ is usually set to large values, depending on the adaptation setup.
In detail, we use $Q=0.9$ for AFE → ExpW and AFE → FER2013, while $Q=0.7$ and $Q=0.8$ for RAF-DB → ExpW and RAF-DB → FER2013.
Our final target model is trained for 50 epochs using SGD, following a cosine annealing scheduling policy.

\subsection{Comparisons}
\label{sec:exp:comparisons}

To the best of our knowledge, \mname~is the first method to tackle SFUDA for FER, therefore we propose a comparison with state-of-the-art methods in the less restrictive UDA setting. To extend the comparison, we also report the results of a couple of general-purpose methods for SFUDA which we re-purposed for FER.
\mname~can be applied seamlessly to an arbitrary SSL method, to this end we report two versions where we apply different self-supervised pretraining on the target domain using SwAV~\cite{caron2020unsupervised} and DeepClusterV2~\cite{caron2018deep,caron2020unsupervised}.

Tab.~\ref{tab:exp:main_results} shows the classification accuracy of competing methods under different domain adaptation settings. Compared among SFUDA methods, our method with SwAV as self-supervised pretraining always performs better than SHOT by over ten points in most of the benchmarks.
The same advantage holds when we adapt from the two source domains to FER2013, with a total improvement of $+4\%$.
More interestingly, \mname\ demonstrates comparable adaptation performance even when compared with UDA methods which have access to the source data. In particular, 
when we adapt from AFE to FER2013, our \mname\ scores the best performance among all methods.

For an in-depth investigation of how SFUDA methods performing on FER, we present the class-wise accuracy when adapting from RAF-DB to FER2013 in Tab.~\ref{tab:exp:classwise_results}. Noticeably, \mname\ manages to consistently adapt better among all classes compared to SHOT and SHOT-IM, where for some classes, e.g. Surprise, Happiness, Sadness, the improvement is greater than $+5\%$. We also notice that for minor classes under the SFUDA setting, e.g. Disgust, the classification accuracy is much lower compared to other major classes, mostly due to the limited samples for expressing the class in the target domain under a large domain gap.

\begin{figure*}[h]
\begin{center}
\begin{tabular}{ccc}
\includegraphics[width=3.4cm]{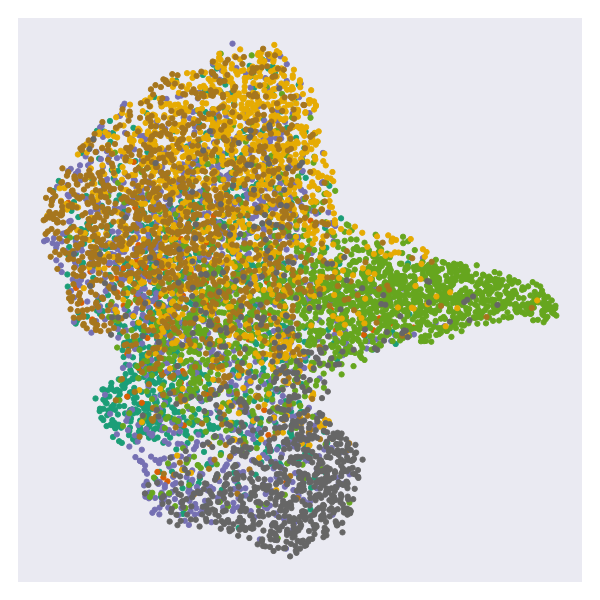} &
\includegraphics[width=3.4cm]{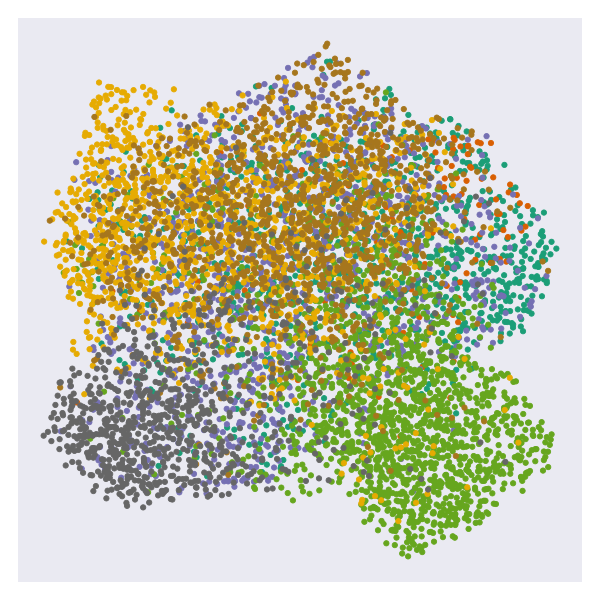} &
\includegraphics[width=3.4cm]{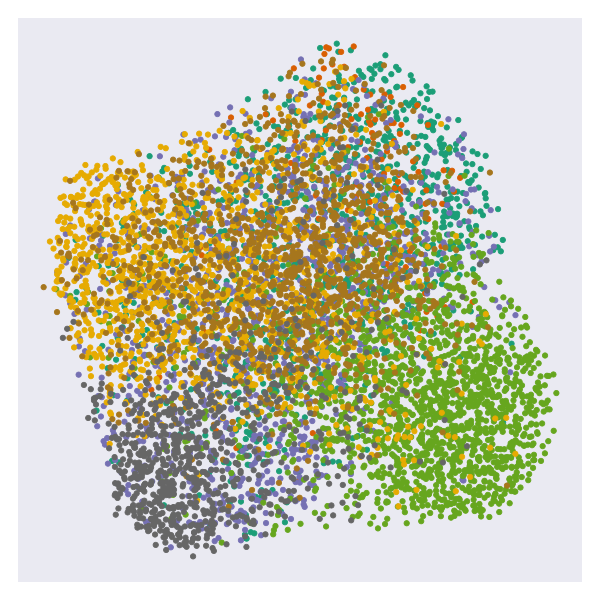} \\
(a) SHOT & (b) DeepClusterV2 (Ours) & (c) SwAV (Ours)
\end{tabular}
\end{center}
\caption{UMAP visualisations of the features spaces in the RAF-DB → FER2013 setting.}
\label{fig:exp:results:umap}
\end{figure*}

\noindent\textbf{Qualitative Result.}
In Fig.~\ref{fig:exp:results:umap}, we present the UMAP visualisation of our methods with the target model pretrained with two self-supervised methods, i.e. DeepClusterV2 and SwAV, in the RAF-DB → FER2013 setting and compare them with SHOT.
SHOT shows a more peculiar shape compared to our models, as it finetunes the source model to the new domain, thus having a tighter relationship with the source data.
The inherited space indirectly constrains the target model to imitate the source domain when moulding the target domain.
Therefore, starting from the source model can hinder the adaptability to the new domain in situations of extreme domain shift.
On the contrary, the UMAPs of our models seem to fit better to the target domain.
While similar, these two embedding spaces present subtle differences.
Visually, the DeepClusterV2 space manages to better separate emotions.

\subsection{Ablation study}
\label{sec:exp:ablation}

We present a thorough analysis of the main design choices of \mname. We first investigate different pretrained networks to validate the effectiveness of the self-supervised pretraining of the target model. We then compare our novel cluster purity score against the traditional confidence score to justify its advantages in providing more reliable pseudo-labels.
Finally, we examine different combinations of our proposed elements and show how they impact the final adaptation performance on FER. 

\noindent\textbf{Does the self-supervised pretrained backbone work better?} In order to understand how each pretrained backbone model serves as the target model, we experiment four options including (i) a model pretrained on ImageNet, (ii) the source model, \ie``Source'', and (iii) the DeepClusterV2 and (iv) the SwAV self-supervised models. For all models, we train a linear classifier applied on top of their frozen feature extractor with ground-truth target labels.
As shown in Tab.~\ref{tab:exp:ablation:backbone}, the self-supervised pretraining on the target domain using SwAV scores the \textit{best} classification accuracy on the two target datasets. DeepClusterV2 demonstrates less consistent improvements over the source model on the two target domain, with $+1.1\%$ improvement on FER2013 dataset, but with $-6.3\%$ on ExpW. This might be due to the superiority of SwAV to learn discriminative feature representations over DeepClusterV2.

\begin{table}[t!]
\begin{center}
\resizebox{0.6\linewidth}{!}{
\begin{tabular}{|ccccc|c|}
\hline
Dataset      & ImageNet & Source  & DeepClusterV2 & SwAV       \\
\hline\hline
ExpW         & 54.45    & 66.80   & 60.54         & \bf{69.13} \\
FER2013      & 36.38    & 56.99   & 58.10         & \bf{60.90} \\
\hline
\end{tabular}
}
\end{center}
\caption{Top-1 accuracy with different pretrained target backbones: a model pretrained on ImageNet, the source model, and two self-supervised models, \ie DeepClusterV2 and SwAV.}
\label{tab:exp:ablation:backbone}
\end{table}

\begin{figure*}[t]
\begin{center}
\begin{tabular}{c}
\includegraphics[width=0.55\textwidth]{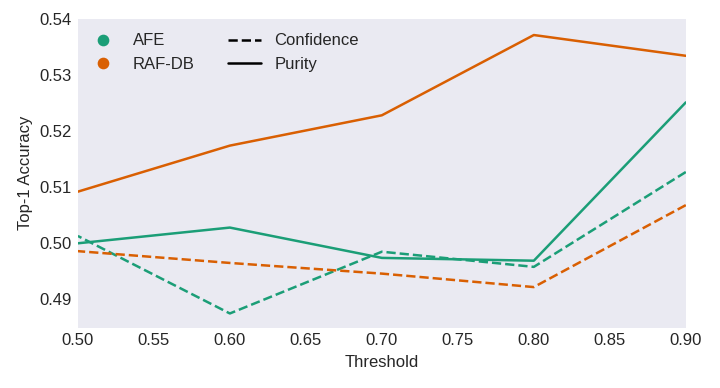}\\
\end{tabular}
\end{center}
   \caption{Top-1 accuracy of \mname\ on FER2013, when adapting from AFE (the green plots) and RAF-DB (the orange plots), with varying thresholds on the confidence (the dashed line) and our cluster purity score (the solid line). Best viewed in colour. 
   }
\label{fig:exp:ablation:conf_vs_purity}
\end{figure*}

\noindent\textbf{Does cluster purity perform better than confidence?}
We ablate our novel cluster purity score in comparison to the traditional confidence to prove its capability of providing more reliable pseudo-labels. We also show the impacts of different threshold on $Q$ ranging from $0.5$ and $0.9$ on the adaptation performance.
Fig.~\ref{fig:exp:ablation:conf_vs_purity} shows the top-1 accuracy of \mname\ on FER2013, when adapting from AFE (the green plots) and RAF-DB (the orange plots), with varying thresholds on the confidence (the dashed line) and our cluster purity score (the solid line). We can observe a general increasing tendency of the accuracy as the threshold value increases, as more reliable pseudo-labels are selected due to a stricter criterion.
Our cluster purity consistently outperforms the confidence at all threshold values. Specifically, when adapting from RAF-DB to FER2013, cluster purity outperforms confidence at the threshold of $80\%$ by more than $+3\%$.

\begin{table*}
\begin{center}
\resizebox{0.95\linewidth}{!}{
\begin{tabular}{|cccccc|c|}
\hline
Backbone & Score     & AFE → ExpW & AFE → FER2013 & RAF-DB → ExpW & RAF-DB → FER2013  \\
\hline\hline
Source   & Conf.  & 56.43      & 48.36         & 59.79        & 50.47            \\
Source   & Purity & 56.54      & 47.34         & 61.18        & \bf{54.29}       \\
SwAV     & Conf.  & 62.88      & 51.27         & 63.22        & 50.68            \\
SwAV     & Purity & \bf{65.00} & \bf{52.51}    & \bf{66.60}   & 53.71            \\
\hline
\end{tabular}}
\end{center}
\caption{Top-1 accuracy of various versions of \mname\ evaluated on four adaptation setups.}
\label{tab:exp:ablation:total}
\end{table*}

\noindent\textbf{How do all the components interact with one another?}
We show how different pretrained target models and different pseudo-label criteria incrementally impact the performance of our proposed method. We consider two backbones, Source and SwAV, and two pseudo-label criteria, confidence and cluster purity score. We present the classification accuracy under different adaptation setups in Tab.~\ref{tab:exp:ablation:total}. Regarding the target model, self-supervised pretraining, \ie SwAV, outperforms the source model under the majority of the adaptation setups, regardless the usage of either confidence or cluster purity score for pseudo-label refinement. When our proposed cluster purity is applied, we observe a consistent improvement of about $+3\%$ over all the adaptation setups with a cluster-based self-supervised feature extractor. When applied on the source model, on the other hand, its advantages are not stable.

%%%%%%%%%
\section{Conclusions}
\label{sec:conclusions}
In this work, we presented the first Source-Free Unsupervised Domain Adaptation solution for Facial Expression Recognition, motivated by the privacy-sensitive nature of facial images. Our method, \mname, employs self-supervised pretraining on the target domain for warming up the target model. To reliably transfer the task knowledge from the source model, \mname~proposes a novel cluster-level pseudo-labelling strategy by refining the pseudo-labels using cluster statistics. We experimentally proved the effectiveness of \mname\ in improving the adaptation performance under various adaptation setups, scoring the new state-of-the-art in terms of FER under the SFUDA setting.
As future work, we aim to extend our method towards online SFUDA, where the adaptation happens as the target data streams in.

\section*{Acknowledgement}
This work was supported by the EU JPI/CH SHIELD project, by the PRIN project PREVUE (Prot. 2017N2RK7K), the EU H2020 MARVEL (957337) project, the EU ISFP PROTECTOR (101034216) project, the EU H2020 SPRING project (871245), and by Fondazione VRT.
It was carried out under the ``Vision and Learning joint Laboratory'' between FBK and UNITN.

\bibliography{bmvc_final.bbl}
\end{document}